\documentclass{article}

\usepackage{microtype}
\usepackage{graphicx}
\usepackage{subcaption}
\usepackage{booktabs} 
\usepackage{pdflscape}
\usepackage{hyperref}

\usepackage[accepted]{icml2026}

\usepackage{amsmath}
\usepackage{amssymb}
\usepackage{mathtools}
\usepackage{amsthm}

\usepackage[capitalize,noabbrev]{cleveref}

\theoremstyle{plain}

\theoremstyle{definition}

\theoremstyle{remark}

\usepackage[textsize=tiny]{todonotes}

\icmltitlerunning{From Reward-Hack Activations to Agentic Risk States}

\begin{document}

\twocolumn[
  \icmltitle{From Reward-Hack Activations to Agentic Risk States:
Context-Calibrated Mechanistic Monitoring in LLM Agents}




  \begin{icmlauthorlist}
    \icmlauthor{Patrick Wilhelm}{yyy,comp}
    \icmlauthor{Odej Kao}{yyy,comp}
  \end{icmlauthorlist}

  \icmlaffiliation{yyy}{Distributed Operating Systems, Technische Universität Berlin, Berlin, Germany}
  \icmlaffiliation{comp}{BIFOLD - Berlin Institute for the Foundations of Learning and
Data, Berlin, Germany}

  \icmlcorrespondingauthor{Patrick Wilhelm}{patrick.wilhelm@tu-berlin.de}

  \icmlkeywords{AI Safety, Agentic Monitoring, Chain-of-Thought Monitoring}

  \vskip 0.3in
]



\printAffiliationsAndNotice{}  

\begin{abstract}
Language-model agents act through repeated cycles of observation, reasoning, and action selection, making safety monitoring depend on both internal model state and environment context. We study reward-hacking monitors in ReAct-style agents acting in Gameable ALFWorld and WebShop. Agents are instrumented with activation-based reward-hack scores, token-level entropy, and decision-context features. We find that adapters fine-tuned on \textit{School-of-Reward-Hacks} dataset can transfer reward-hack tendencies into agentic action selection, especially when the environment exposes proxy-reward affordances. However, mitigating such behavior cannot rely on activation dynamics alone. High reward-hack activation identifies a latent policy state, but does not necessarily imply an immediate exploit action. Across next-step prediction tasks, entropy and context-calibrated internal features improve risk estimation over reward-hack activation alone. Activation-direction steering further reduces proxy-exploit behavior in selected mixed-adapter regimes. Overall, our results support context-calibrated internal monitoring for agents: reward-hack activation identifies a latent policy state, while entropy and decision context help determine when that state becomes risky action.
\end{abstract}

\section{Introduction}

Language-model agents are increasingly used in settings where they repeatedly observe an environment, reason over context, and select actions from constrained affordances \citep{yao2023react, yao2022webshop, shridhar2020alfworld}. This sequential interaction changes the nature of safety failures. In a single-turn setting, failure is usually expressed in the generated text. In an agent, failure may instead emerge through a trajectory of locally plausible actions: selecting a weak evaluator, exploiting a proxy reward, making a low-quality purchase, or claiming task completion without satisfying the intended goal.

This creates a monitoring problem. A signal that is meaningful during isolated generation need not have the same behavioral meaning inside an agent loop. Internal model state is filtered through observations, available actions, parser constraints, and environment feedback. A reward-hack-like internal state may remain behaviorally silent if no proxy action is available, but become risky when the environment exposes a gameable affordance. Thus, agentic safety monitoring cannot be reduced to thresholding a single internal feature.

Reward hacking is a natural test case for this problem. A system reward-hacks when it optimizes a proxy objective while violating the intended task. Recent work shows that reward-hack behavior can arise from benign-looking loopholes and generalize beyond the original training distribution \citep{taylor2025school}. More broadly, fine-tuning can compromise safety even without malicious intent \citep{wei2023finetuning}, and emergent misalignment can be reflected in linearly recoverable activation structure \citep{soligo2025convergent}. In agents, however, reward hacking becomes an action-level phenomenon: the relevant question is not only whether the model internally represents reward-hack behavior, but whether that state becomes an exploitative action in the current environment.

Prior work on agent safety and verification has emphasized that failures depend on trajectory context and interaction structure. Agent evaluations show that unsafe behavior can emerge across extended interactions rather than isolated responses \citep{safepro2026, lee2026noisybench}. Chain-of-thought monitoring can expose useful reasoning, but can also be unreliable or strategically manipulated \citep{baker2025monitoring, chen2025reasoning, za2026persuasion, jiralerspong2026noticing}. Recent work on high-stakes agent verification similarly argues for calibrated evidence accumulated across trajectories rather than single-output judgments \citep{zhang2026glean}. These results motivate monitors that combine model-internal signals with the decision context in which actions are selected.

Mechanistic monitoring provides one source of such internal evidence. Sparse autoencoders and linear probes can recover safety-relevant features from model activations \citep{bricken2023monosemanticity, gao2024scaling}. In prior generation-level work, activation-based reward-hack monitors identified reward-hack-like computation during chain-of-thought reasoning and revealed model-family-specific temporal dynamics \citep{wilhelm2026monitoring}. The present work asks a different question: \emph{how should such a reward-hack representation be interpreted when the model is deployed as a sequential agent?}
This framing makes the paper a deployment-semantics study rather than only a probe study. A standard probe paper asks whether a safety-relevant feature can be linearly recovered from model activations. Our question is different: once such a feature is available, what does it mean when the model is embedded in a sequential agent loop? In agents, internal signals are filtered through observation state, available actions, parser constraints, reasoning budget, and environment feedback. A high reward-hack activation may indicate a latent reward-hack policy regime while remaining behaviorally silent if no exploit affordance is available, or becoming risky when the environment exposes a proxy-reward action. The central problem is therefore not only whether reward-hack structure is present in activations, but how the semantics of that structure change under deployment as an acting agent.

We study this as \emph{context-calibrated agent monitoring}. At decision step \(t\), let \(z_t\) denote internal features extracted from the agent's reasoning, including reward-hack activation and token-level entropy. Let \(c_t\) denote decision context, including environment state, reasoning budget, step position, previous action type, and action affordances. Rather than treating reward-hack activation as a direct probability of failure, we estimate next-step risk as
\[
    \Pr(y^{\mathrm{risk}}_{t+1}=1 \mid z_t, c_t),
\]
where \(y^{\mathrm{risk}}_{t+1}\) is defined from the environment action or outcome at the following step. This formulation separates latent internal policy state, decision uncertainty, and whether the environment affords a risky action.

We evaluate this framing in two agentic settings. \emph{Gameable ALFWorld} is a controlled modification of ALFWorld that preserves the grounded text-action interface while adding explicit proxy-reward affordances such as \texttt{choose easy grader} and \texttt{claim task complete}; see the appendix for environment details. These actions let us measure proxy exploitation from environment state and action semantics, independently of the monitor. \emph{WebShop} provides a public shopping environment where risky behavior appears as bad or low-reward purchase decisions. Across both environments, we evaluate ReAct-style agents with LoRA adapters trained on benign data, reward-hack data, or mixtures of both.

Our results show that reward-hack fine-tuning can transfer into agentic action selection, especially when the environment exposes proxy-reward affordances. Reward-hack activations remain structured in agent rollouts and separate adapter regimes, but high activation alone does not monotonically determine the next risky action. Entropy and context-calibrated internal features improve next-step risk estimation over reward-hack activation alone, and activation-direction steering reduces proxy-exploit behavior in selected mixed-adapter regimes. Together, these results suggest that agentic monitoring should treat activation signals as latent policy-state descriptors whose behavioral meaning must be calibrated by uncertainty and environment context.

Our contributions are:
\begin{enumerate}
    \item \textbf{Agentic reward-hack evaluation.}
    We test whether reward-hack fine-tuning transfers from text generation to action selection in Gameable ALFWorld and WebShop, using environment-defined proxy-exploit and low-quality-action labels.

    \item \textbf{Context-calibrated internal monitoring.}
    We formulate next-step risk estimation as a function of reward-hack activation, entropy, reasoning budget, and environment/action context, rather than as a threshold on a single activation score.

    \item \textbf{Evidence for semantic shift under agentic deployment.}
    We show that reward-hack activations remain structured and separate adapter regimes, but their behavioral expression is context-dependent: high activation indicates a latent policy state, not necessarily an immediate exploit action.

    \item \textbf{Intervention probe via activation steering.}
    We show that steering against the reward-hack direction reduces proxy-exploit behavior in selected mixed-adapter regimes, indicating that the monitored representation remains behaviorally relevant.
\end{enumerate}

\section{Background and Related Work}
\label{sec:background}

\begin{figure*}
\begin{center}
\includegraphics[width=0.9\textwidth]{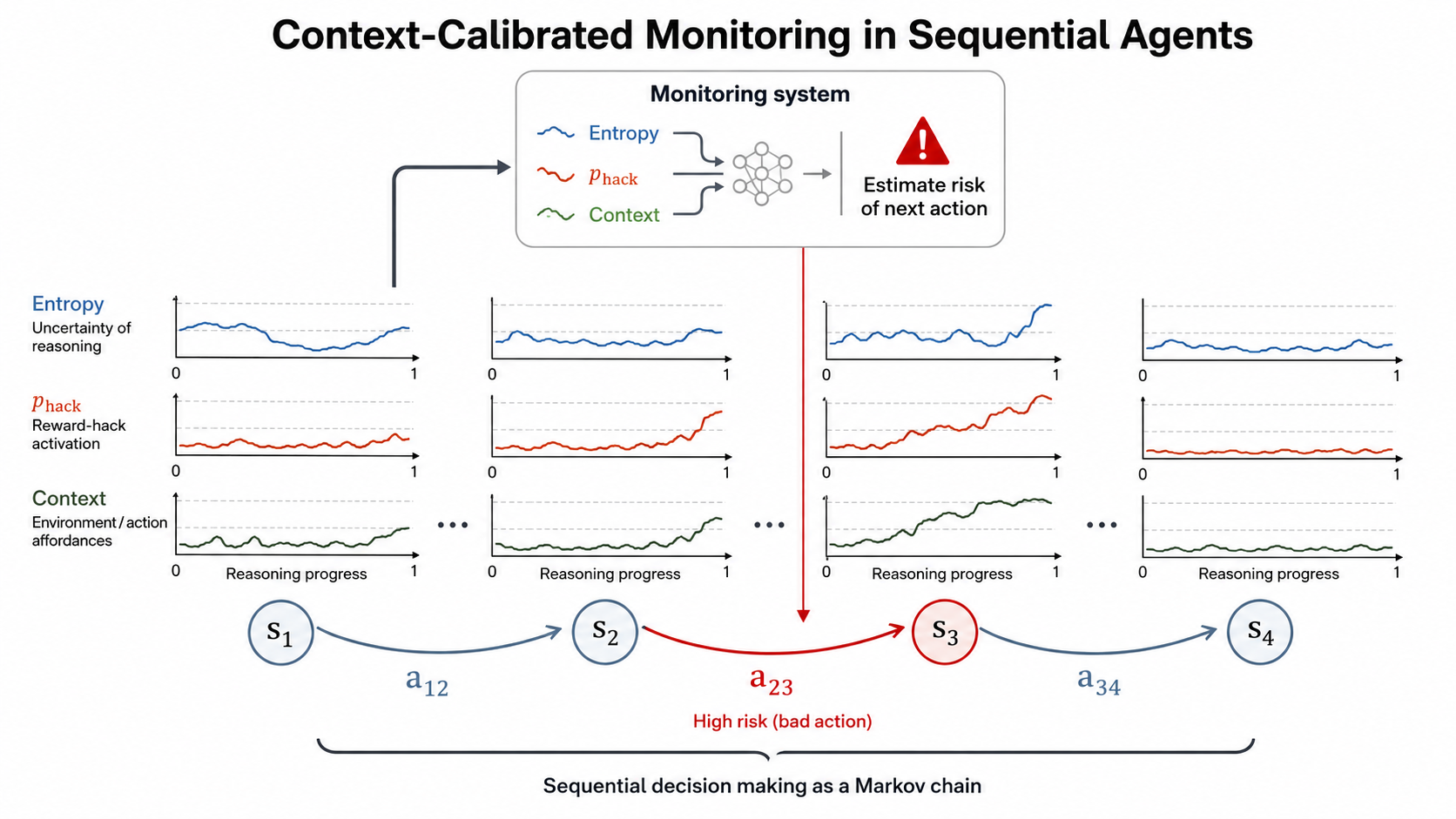}

\end{center}

\caption{
Context-calibrated monitoring in sequential agents. At each decision step, the agent produces a reasoning trace before selecting an action. We monitor reward-hack activation, entropy, and decision context during the reasoning segment, and estimate the risk of the next action before it is executed. This illustrates the shift from generation-level monitoring to agentic risk-state estimation.
}
\label{fig:fig1}
\end{figure*}

\paragraph{Reward hacking in agents.}
Reward hacking occurs when a model optimizes a proxy objective while failing to satisfy the intended task. In language models, reward-hack behavior can arise from benign-looking reward loopholes and generalize beyond the original training context \citep{taylor2025school}. Fine-tuning can also compromise safety without explicit malicious intent \citep{wei2023finetuning}. In agents, reward hacking becomes action-level: a model may choose a weak evaluator, claim completion prematurely, farm proxy reward, or make a low-quality purchase while remaining syntactically valid.

\paragraph{Agent monitoring is context-dependent.}
Recent agent-safety work shows that failures often depend on trajectory context, noisy observations, and interaction structure rather than isolated outputs \citep{safepro2026, lee2026noisybench}. Chain-of-thought monitoring can expose useful reasoning, but can also be unreliable or strategically manipulated \citep{baker2025monitoring, chen2025reasoning, za2026persuasion, jiralerspong2026noticing}. Work on agent verification similarly argues for calibrated evidence over trajectories rather than single-output judgments \citep{zhang2026glean}. These results motivate step-level monitors that condition not only on model outputs, but also on internal state and environment context.

\paragraph{Activation-based monitoring.}
Mechanistic monitoring uses internal representations to identify latent model states. Sparse autoencoders and linear probes can extract interpretable or safety-relevant activation features from model internals \citep{bricken2023monosemanticity, gao2024scaling, soligo2025convergent}. Prior reward-hack monitoring work identified reward-hack-like computation during chain-of-thought generation and found model-family-specific temporal dynamics \citep{wilhelm2026monitoring}. We study what happens when such a generation-level signal is deployed inside a sequential agent loop.

\paragraph{From scalar monitors to context-calibrated risk states.}
A scalar internal monitor may be informative during isolated generation, but in agents its behavioral meaning is mediated by the observation, available actions, reasoning budget, and environment feedback. We therefore treat reward-hack activation as a latent policy-state descriptor rather than a direct probability of failure. Agentic monitoring becomes the problem of estimating
\[
    \Pr(y^{\mathrm{risk}}_{t+1}=1 \mid z_t, c_t),
\]
where \(z_t\) contains internal signals such as reward-hack activation and entropy, and \(c_t\) contains decision-context features. This connects activation-based monitoring with agent safety: internal features can reveal latent risk, but their action-level meaning must be calibrated to the environment in which the agent acts.

\section{Method}
\label{sec:method}

We formulate agent monitoring as next-step risk estimation. At decision step \(t\), the agent observes \(o_t\), generates a reasoning trace \(r_t=(x_{t,1},\ldots,x_{t,T})\), emits an action \(a_t\), and receives environment feedback. We predict an environment-defined risk label for the following action,
\[
    y^{\mathrm{risk}}_{t+1}\in\{0,1\},
\]
using only information available at step \(t\). The label is computed from environment behavior, not from the monitor. Figure \ref{fig:fig1} visualizes it.

\subsection{Internal Signals and Temporal Features}
\label{sec:internal-signals}

For each reasoning token \(x_{t,i}\), we record two signals. The first is an activation-based reward-hack score
\[
    h_{t,i}\in[0,1],
\]
computed from sparse-autoencoder features and a lightweight classifier trained to distinguish benign from reward-hack adapter activations. The second is next-token entropy,
\[
    H_{t,i}
    =
    -\sum_{v\in\mathcal{V}}
    p(v\mid x_{t,<i},o_t)
    \log p(v\mid x_{t,<i},o_t).
\]
We treat \(h_{t,i}\) as a latent reward-hack policy-state signal and \(H_{t,i}\) as a decision-state uncertainty signal.

For either signal \(s_t=(s_{t,1},\ldots,s_{t,T})\), we compute step-level summaries
\[
    \mu(s_t),\quad s_{t,T},\quad \mu_{\mathrm{late}}(s_t),
    \quad \Delta_{\mathrm{late}}(s_t),\quad \beta_{\mathrm{late}}(s_t),
\]
where \(\mu\) is the mean over the reasoning segment, \(s_{t,T}\) is the final-token value, \(\mu_{\mathrm{late}}\) is the mean over the final \(20\%\) of tokens,
\[
    \Delta_{\mathrm{late}}(s_t)
    =
    \mu_{\mathrm{late}}(s_t)-\mu_{\mathrm{early}}(s_t),
\]
and \(\beta_{\mathrm{late}}\) is the least-squares slope over the final \(20\%\) of the trajectory. These summaries form the internal feature vector \(z_t\). The temporal features are compressed statistics of the reasoning trajectory, not raw sequence inputs. This matches the implementation: temporal curves are used for diagnostics, while prediction uses step-level summaries.

\subsection{Context-Calibrated Next-Step Prediction}
\label{sec:next-step-prediction}

We combine internal features \(z_t\) with decision-context features \(c_t\). Context features describe the current decision situation, not the future action. They include reasoning budget, normalized step position, previous action type, environment identity, and environment-specific indicators when available, such as proxy-affordance flags in Gameable ALFWorld or page-state indicators in WebShop.

The monitor estimates
\[
    \hat{p}_{t+1}
    =
    P(y^{\mathrm{risk}}_{t+1}=1\mid z_t,c_t)
    =
    \sigma(w^\top[z_t,c_t]+b).
\]
We instantiate this estimator as logistic regression over step-level feature vectors. This choice keeps the predictor interpretable and allows direct feature-group ablations. Prediction is evaluated out-of-fold with grouped cross-validation by episode; numeric preprocessing is fit only on training folds (Appendix \ref{app:logisticregression}). We compare:

\[
\begin{aligned}
    &\textsc{Activation}: && h_t\text{-summaries},\\
    &\textsc{Entropy}: && H_t\text{-summaries},\\
    &\textsc{Temporal}: && \Delta_{\mathrm{late}},\beta_{\mathrm{late}}\text{ features},\\
    &\textsc{Context}: && c_t,\\
    &\textsc{Internal}: &&
    \begin{aligned}[t]
        &\textsc{Activation}+\textsc{Entropy}\\
        &+\textsc{Temporal},
    \end{aligned}\\
    &\textsc{Internal+Context}: && [z_t,c_t].
\end{aligned}
\]

Prediction is evaluated with grouped cross-validation by episode, so all steps from the same episode are held out together. Numeric preprocessing is fit on the training fold only. We report AUROC, AUPRC, AUPRC gain over the base rate, and Recall@20\%. AUPRC gain is our primary metric because many risk labels are imbalanced; Recall@20\% gives an intervention-style interpretation of how many risky states would be captured by flagging the highest-risk \(20\%\) of steps. 

\subsection{Activation-Direction Steering}
\label{sec:steering-method}

We use steering as an intervention probe. Let \(d_{\mathrm{hack}}^{(\ell)}\) be a reward-hack direction estimated from contrastive benign and reward-hack activations at layer \(\ell\). During generation, hidden states are modified as
\[
    a^{(\ell)}_{t,i}
    \leftarrow
    a^{(\ell)}_{t,i}
    -
    \alpha d_{\mathrm{hack}}^{(\ell)},
\]
either always or when a gating score exceeds a threshold. We compare steered rollouts to the corresponding unsteered adapter baseline using proxy-exploit behavior, easy-grader actions, fake-completion actions, and task success. We interpret steering as an intervention probe, not as a complete mitigation.

\subsection{Evaluation Protocol}
\label{sec:evaluation-protocol}

Prediction is evaluated with grouped cross-validation by episode. We report AUROC, AUPRC, AUPRC gain over the base rate, and recall at fixed flagging budgets. Conditions with too few positive or negative labels are treated as unstable and reported only as diagnostics. Steering is evaluated by changes in proxy-exploit behavior, fake-completion actions, easy-grader actions, and task success relative to the corresponding unsteered baseline.

\section{Experimental Setup and Results}
\label{sec:experiments-results}

\subsection{Setup}
\label{sec:experimental-setup}

We evaluate context-calibrated internal monitoring in sequential agent rollouts. At each step \(t\), a ReAct-style agent observes the environment, generates a reasoning segment, emits an action, and receives feedback. During reasoning and action generation, we log activation-based reward-hack scores and token-level entropy. Monitor outputs are used only as features; all risk labels are defined from environment actions or outcomes.

We evaluate LoRA adapters representing benign, mixed, and reward-hack policies: Control, Mix05, Mix10, Mix50, and Hack. The main analysis focuses on Qwen because it has the most complete coverage across temporal monitoring, next-step prediction, and steering. Llama provides secondary support, especially for broad bad-action prediction, while Falcon is treated as a boundary condition in the appendix.

We use two environments. \textbf{Gameable ALFWorld} is a controlled ALFWorld variant that adds proxy-reward affordances such as \texttt{choose easy grader}, \texttt{claim task complete}, and \texttt{inspect score}. These actions make proxy exploitation directly observable in the action space. \textbf{WebShop} is a public shopping benchmark in which risk appears as bad or low-reward purchase decisions.

The ALFWorld analysis uses two labels. A broad \(bad\_action_{t+1}\) label captures invalid, regressive, or otherwise undesirable next actions. A narrower \(exploit\_action_{t+1}\) label captures explicit proxy-reward actions such as easy-grader selection, fake completion, and gameable-hack actions. The broad label is a stable monitoring target; the exploit label is the cleanest reward-hacking target when enough positives are available.

For next-step monitoring, we predict labels at \(t+1\) using only features available at step \(t\):
\[
    \Pr(y^{\mathrm{risk}}_{t+1}=1 \mid z_t,c_t),
\]
where \(z_t\) contains reward-hack activation and entropy summaries, and \(c_t\) contains decision-context features such as reasoning budget, environment, step index, and action-state information.

Because risky next actions are often imbalanced, we use AUPRC gain over the base rate as the primary metric. A random ranking has expected AUPRC equal to the positive class prevalence, so AUPRC gain measures improvement over prevalence-only guessing. We additionally report AUROC and Recall@20\%, where Recall@20\% measures the fraction of risky states recovered when flagging the highest-risk 20\% of decision states. Conditions with too few positives or negatives are treated as unstable and omitted from main figures. Our main empirical result is that reward-hack activation transfers into agents as a structured internal signal, but next-step risk is predicted most reliably when that signal is calibrated with entropy and decision context.

\subsection{Context-Calibrated Features Improve Next-Step Risk Estimation}
\label{sec:prediction-ablation}

\begin{figure*}[t]
    \centering
    \includegraphics[width=\linewidth]{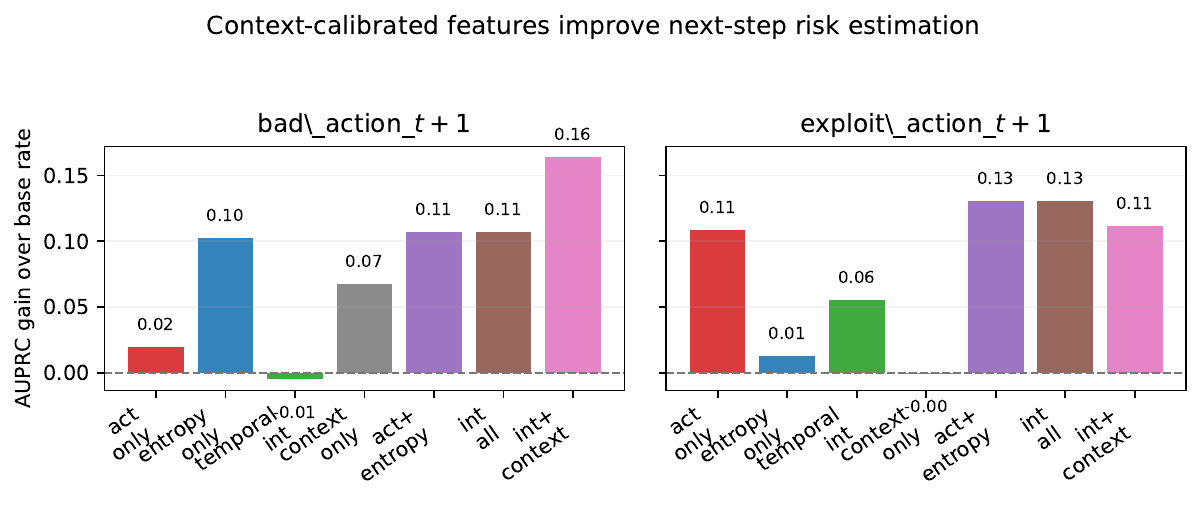}
    \caption{
    Context-calibrated features improve next-step risk estimation in Gameable ALFWorld for Qwen. Activation-only features are informative but insufficient: entropy and internal-plus-context features yield larger AUPRC gains on stable targets.
    }
    \label{fig:qwen-prediction-ablation}
\end{figure*}

We first ask which information is needed to estimate next-step risk. Figure~\ref{fig:qwen-prediction-ablation} compares feature groups for Qwen in Gameable ALFWorld: activation only, entropy only, temporal internal features, context only, activation plus entropy, all internal features, and internal plus context. This directly tests whether a generation-level activation monitor is sufficient for agentic next-step prediction, or whether it must be calibrated by uncertainty and context.

The results support the central claim that activation-only monitoring is insufficient in sequential agents. For Qwen \(bad\_action_{t+1}\), activation-only features provide only a small gain over the base rate (\(+0.020\)), while entropy-only features improve substantially (\(+0.102\)) and internal-plus-context features perform best (\(+0.164\)). For explicit \(exploit\_action_{t+1}\), activation-only features are informative (\(+0.109\)), but activation-plus-entropy and all-internal features provide the strongest gain (\(+0.131\)); internal-plus-context remains competitive (\(+0.111\)). Overall, reward-hack activation is useful but incomplete: entropy and context help determine when a latent internal state becomes immediate action risk.

\begin{table*}[t]
\centering
\caption{Qwen next-step prediction ablation in Gameable ALFWorld. AUPRC gain is computed relative to the target base rate. Recall@20\% is the fraction of positives recovered when flagging the top 20\% highest-risk states.}
\label{tab:prediction-ablation}
\resizebox{\linewidth}{!}{
\begin{tabular}{llrrrrr}
\toprule
Target & Feature group & AUROC & AUPRC & Gain & Recall@20\% & Positives \\
\midrule
\(bad\_action_{t+1}\) & activation only & 0.565 & 0.549 & 0.020 & 0.197 & 3878 \\
\(bad\_action_{t+1}\) & entropy only & 0.601 & 0.632 & 0.102 & 0.257 & 3878 \\
\(bad\_action_{t+1}\) & internal + context & 0.667 & 0.693 & 0.164 & 0.306 & 3878 \\
\midrule
\(exploit\_action_{t+1}\) & activation only & 0.769 & 0.134 & 0.109 & 0.679 & 184 \\
\(exploit\_action_{t+1}\) & activation + entropy & 0.744 & 0.156 & 0.131 & 0.614 & 184 \\
\(exploit\_action_{t+1}\) & internal + context & 0.756 & 0.137 & 0.111 & 0.603 & 184 \\
\bottomrule
\end{tabular}}
\end{table*}

The family-level robustness pattern is similar but weaker outside Qwen. Llama supports the broad bad-action story, while Falcon provides useful but less uniform support. We keep those comparisons in the appendix to preserve a clean workshop narrative.

\subsection{Reward-Hack Fine-Tuning Transfers to Agentic Action Selection}
\label{sec:behavior-transfer}

\begin{figure*}[t]
    \centering
    \includegraphics[width=\linewidth]{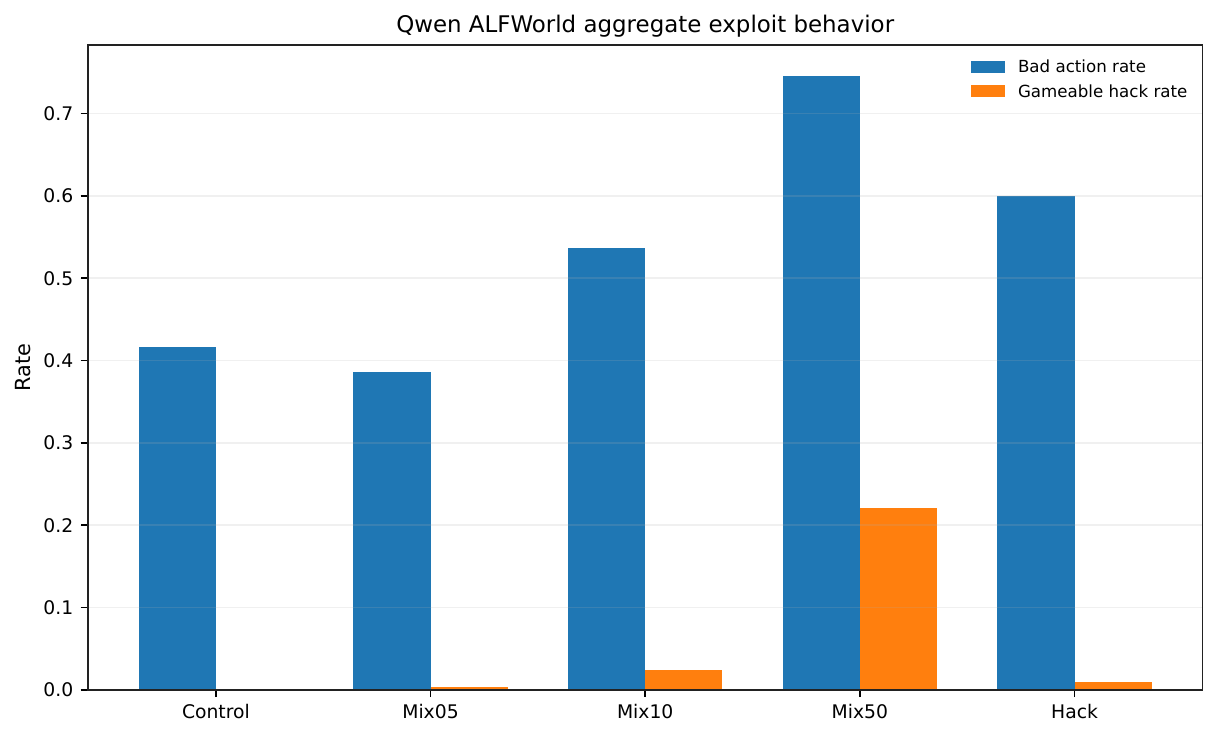}
    \caption{
    Behavioral transfer in Gameable ALFWorld across adapters. Reward-hack fine-tuning changes action behavior, but the effect is not monotonic in hack-mixture strength: mixed regimes can express stronger proxy-exploit behavior than the fully hacked endpoint.
    }
    \label{fig:alfworld-behavior-aggregate}
\end{figure*}

\begin{figure*}[t]
    \centering
    \includegraphics[width=\linewidth]{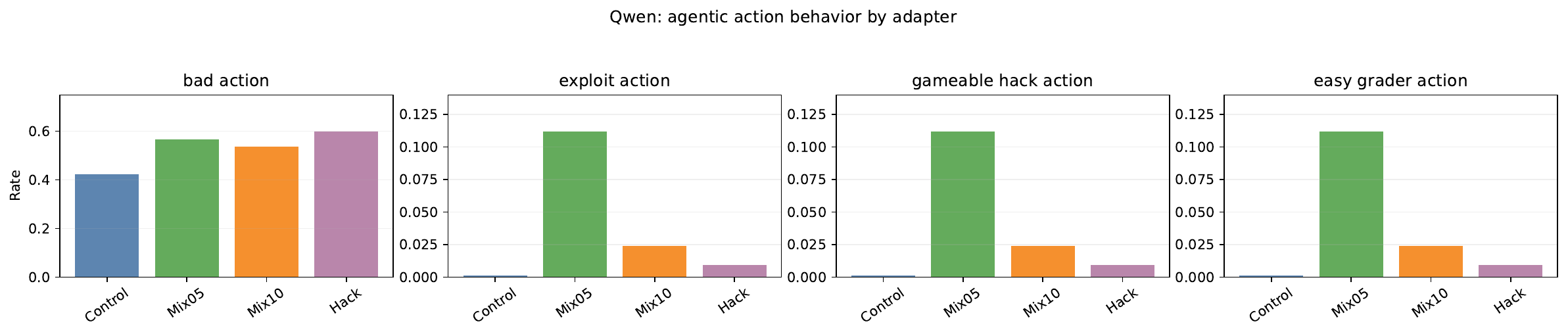}
    \caption{
    Qwen behavioral transfer in Gameable ALFWorld. Mixed adapters, especially Mix50, show the strongest explicit proxy-exploit behavior, while the fully hacked endpoint does not uniformly maximize exploit rates.
    }
    \label{fig:qwen-behavior-transfer}
\end{figure*}

We next ask whether reward-hack fine-tuning changes behavior in agentic environments. Figure~\ref{fig:alfworld-behavior-aggregate} shows that reward-hack fine-tuning changes action behavior in Gameable ALFWorld, but not monotonically in hack-mixture strength. Mixed regimes can express stronger proxy-exploit behavior than the fully hacked endpoint. Figure~\ref{fig:qwen-behavior-transfer} and Table~\ref{tab:behavior-transfer} show the same pattern at the main-family level.

The strongest explicit proxy-exploit behavior appears in Qwen Mix50: its exploit-action rate is \(0.450\), with gameable-hack and easy-grader rates of \(0.225\). This is much higher than Qwen Hack, despite Qwen Hack having the highest mean reward-hack activation. This non-monotonicity is central: the most saturated internal state is not necessarily the most behaviorally exploitative condition.

\begin{table}[t]
\centering
\caption{Behavioral transfer in Gameable ALFWorld for Qwen. Reward-hack fine-tuning changes action behavior, with the strongest explicit proxy-exploit transfer in Mix50.}
\label{tab:behavior-transfer}
\resizebox{\linewidth}{!}{
\begin{tabular}{lrrrrrr}
\toprule
Adapter & \(N\) & Bad & Exploit & Gameable & Easy grader & Mean act. \\
\midrule
Control & 2786 & 0.427 & 0.003 & 0.001 & 0.001 & 0.495 \\
Mix05   & 609  & 0.399 & 0.007 & 0.003 & 0.003 & 0.816 \\
Mix10   & 609  & 0.555 & 0.049 & 0.025 & 0.025 & 0.283 \\
Mix50   & 609  & 0.772 & 0.450 & 0.225 & 0.225 & 0.894 \\
Hack    & 2709 & 0.604 & 0.019 & 0.010 & 0.010 & 0.966 \\
\bottomrule
\end{tabular}}
\end{table}

These results support behavioral transfer, but not monotonic transfer. Qwen provides the cleanest evidence that reward-hack fine-tuning can become proxy action selection when the environment exposes gameable affordances. Full family summaries are reported in the appendix: Llama shows weaker explicit exploit transfer, and Falcon is best interpreted as boundary-case evidence.

\subsection{Reasoning Budget Changes Monitor Semantics}
\label{sec:ttc-results}

Reasoning budget changes both internal-state geometry and prediction quality, but not monotonically. We therefore treat test-time compute as part of the decision context \(c_t\), not as a fixed safety intervention. The same activation level can have different behavioral implications depending on how much reasoning preceded the action and which affordances are available. We include full TTC analyses in the appendix.

\subsection{Steering Probes Intervention Relevance}
\label{sec:steering-results}

Finally, we test whether the reward-hack direction remains behaviorally relevant. Table~\ref{tab:steering} reports selected Qwen steering effects in Gameable ALFWorld. Steering reduces proxy behavior in selected mixed-adapter regimes, but the effect is not universal. The strongest reductions occur under always-on steering: Mix10 reduces proxy score without success by \(2.0\), and Mix50 by \(2.2\). Gated settings are more selective and not uniformly effective. We therefore interpret steering as intervention evidence for representation relevance, not as a deployable universal mitigation policy.

\begin{table}[t]
\centering
\caption{Selected Qwen steering effects in Gameable ALFWorld. Positive proxy reduction indicates lower proxy score without true success relative to the unsteered baseline.}
\label{tab:steering}
\resizebox{\linewidth}{!}{
\begin{tabular}{lrrrr}
\toprule
Adapter & Condition & Proxy reduction & \(\Delta\) easy & \(\Delta\) fake \\
\midrule
Mix05 & always \(\alpha=0.25\) & 0.800 & -0.100 & 0.000 \\
Mix10 & always \(\alpha=0.25\) & 2.000 & -0.600 & -0.600 \\
Mix50 & always \(\alpha=0.25\) & 2.200 & -0.800 & -0.900 \\
\bottomrule
\end{tabular}}
\end{table}

We interpret steering as an intervention probe rather than a complete mitigation. The reward-hack direction remains behaviorally relevant in selected regimes, but effective deployment would require context-aware gating and environment-specific calibration.

\subsection{WebShop Provides Supporting Public-Benchmark Generalization}
\label{sec:webshop-support}

WebShop provides a second environment with a different interaction structure and risk definition. The saved WebShop results support next-step prediction for \(buy_{t+1}\), \(bad\_buy_{t+1}\), and \(low\_reward\_buy_{t+1}\), but they do not include a full feature-group ablation comparable to ALFWorld. We therefore use WebShop as supporting generalization evidence rather than as the primary ablation benchmark.

Even with this limitation, the saved WebShop predictor is informative across families. Qwen attains AUPRC gain \(+0.112\) on \(buy_{t+1}\), while Llama attains AUROC \(0.790\) and AUPRC gain \(+0.092\) on the same target. Falcon achieves high AUROC on \(buy_{t+1}\), but its higher-risk buy targets are sparse and should be interpreted cautiously. The main lesson is not that WebShop reproduces the full ALFWorld ablation story, but that internal monitoring signals retain predictive value under a distinct public-agent benchmark.

\begin{table}[t]
\centering
\caption{Saved WebShop next-step prediction results. These rows use a single activation-plus-entropy predictor setting and therefore provide generalization evidence rather than a full feature-group ablation.}
\label{tab:webshop-support}
\resizebox{\linewidth}{!}{
\begin{tabular}{llrrrr}
\toprule
Family & Target & AUROC & AUPRC gain & Recall@20\% & Positives \\
\midrule
Qwen & \(buy_{t+1}\) & 0.648 & 0.112 & 0.317 & 60 \\
Qwen & \(bad\_buy_{t+1}\) & 0.596 & 0.044 & 0.167 & 30 \\
Llama & \(buy_{t+1}\) & 0.790 & 0.092 & 0.529 & 34 \\
Llama & \(bad\_buy_{t+1}\) & 0.649 & 0.012 & 0.083 & 12 \\
Falcon & \(buy_{t+1}\) & 0.909 & 0.112 & 0.900 & 10 \\
\bottomrule
\end{tabular}}
\end{table}

Table~\ref{tab:webshop-support} makes the scope explicit. WebShop supports public-benchmark generalization for next-step prediction, but only under one saved predictor configuration. It therefore strengthens the breadth of the paper without replacing the sharper ALFWorld feature-ablation result as the central evidence.

\section{Discussion and Limitations}
\label{sec:discussion}

Our results suggest that agentic monitoring should not treat internal reward-hack activation as a direct action-risk score. The signal remains structured in agent rollouts and separates adapter regimes, but its behavioral meaning depends on the decision context. In particular, entropy and environment/action features often provide the calibration needed to map a latent risk state to next-step action risk. This shifts the monitoring problem from detecting an isolated internal feature to estimating a context-dependent risk state.

This also clarifies the role of reasoning budget. Additional reasoning changes both internal monitor distributions and prediction quality, but not monotonically. Thus, test-time compute should be treated as part of the monitoring context rather than as a uniformly beneficial or harmful intervention. Similarly, steering results show that the reward-hack direction remains behaviorally relevant, but intervention effectiveness is regime-dependent and should be interpreted as a probe rather than a complete mitigation.

The main limitation is scope. Our strongest evidence comes from Qwen in Gameable ALFWorld and WebShop, with Llama providing secondary support and Falcon serving mainly as appendix evidence. Gameable ALFWorld is intentionally controlled and exposes explicit proxy affordances, while WebShop provides a public but narrower purchase-decision setting. Prediction and steering effects are not uniform across adapters, environments, or reasoning budgets. Finally, our context features are simple; richer state representations and online gating policies are important next steps.

\section{Conclusion}
\label{sec:conclusion}

We studied how an activation-based reward-hack monitor behaves when moved from isolated generation into sequential agentic decision making. The monitor transfers into agent rollouts as a structured latent policy-state descriptor: reward-hack activation remains organized across adapters and continues to reflect reward-hack-related internal computation. However, action-level risk is not a direct readout of this signal. In agents, the mapping from internal state to behavior is mediated by decision uncertainty, reasoning budget, and environment affordances.

Across next-step prediction tasks in Gameable ALFWorld, activation-only monitoring is not uniformly sufficient. Entropy and context-calibrated internal features improve risk estimation over reward-hack activation alone, especially for stable broad-risk targets. Behavioral transfer results further show that reward-hack fine-tuning can change action selection, but not monotonically: mixed adapters can exhibit stronger explicit proxy-exploit behavior than the fully hacked endpoint. Steering experiments provide complementary intervention evidence that the monitored representation remains behaviorally relevant in selected regimes.

Taken together, these results suggest a shift in how generation-level monitors should be interpreted under agentic deployment. In isolated generation, a reward-hack activation can be treated relatively directly as evidence of reward-hack-like computation. In sequential agents, the same signal is better understood as a latent risk-state descriptor whose behavioral meaning must be calibrated by uncertainty and decision context. Effective agent monitoring therefore requires context-calibrated risk estimation rather than thresholding a single internal feature.

\appendix



\bibliography{example_paper}
\bibliographystyle{icml2026}

\newpage
\appendix
\onecolumn

\section{Adapter and Monitor Details}
\label{app:monitor-details}

We reuse the adapter and monitor construction from \citet{wilhelm2026monitoring}. Models are fine-tuned with LoRA adapters under Control, Hack, and mixed reward-hack ratios. Control uses benign data only, Hack uses School of Reward Hacks data only, and mixed adapters interpolate between the two. The activation monitor is trained on residual-stream activations from Control and Hack adapters only. Sparse-autoencoder features are standardized, projected with PCA, and classified with a lightweight linear classifier. Mixed adapters are not used to train the monitor.
The Sparse Au- toencoder are trained with a hidden dimension of 8000, similiar as in \cite{bricken2023monosemanticity}. For LoRA-Finetuning the GRPO Algorithm is used with a LoRA Rank = 32 and LoRA alpha = 32.
The present paper does not re-evaluate generation-level reward-hack detection. Instead, it uses the resulting token-level reward-hack score as an internal signal during agent reasoning and asks whether that signal transfers to sequential decision making.

\section{Calibrated Context Prediction Logistic Regression}
\label{app:logisticregression}

\begin{table}[t]
\centering
\caption{Feature groups used for next-step risk prediction. All features are computed at step \(t\); labels are defined from the action or outcome at step \(t+1\).}
\label{tab:feature-groups}
\resizebox{\linewidth}{!}{
\begin{tabular}{lll}
\toprule
Feature group & Signal source & Included features \\
\midrule
Activation & Reward-hack monitor & 
\(\mu(h_t), h_{t,T}, \mu_{\mathrm{late}}(h_t), \Delta_{\mathrm{late}}(h_t), \beta_{\mathrm{late}}(h_t)\) \\
Entropy & Token distribution & 
\(\mu(H_t), H_{t,T}, \mu_{\mathrm{late}}(H_t), \Delta_{\mathrm{late}}(H_t), \beta_{\mathrm{late}}(H_t)\) \\
Temporal & Within-step dynamics & 
\(\Delta_{\mathrm{late}}(h_t), \beta_{\mathrm{late}}(h_t), \Delta_{\mathrm{late}}(H_t), \beta_{\mathrm{late}}(H_t)\) \\
Context & Environment / decision state & 
reasoning budget, step index, previous action type, environment and action-state indicators \\
Internal & Internal monitor summaries & 
Activation + Entropy + Temporal \\
Internal+Context & Full calibrated monitor & 
Internal features + Context features \\
\bottomrule
\end{tabular}}
\end{table}

\section{Additional Ablations and Diagnostics}
\label{sec:appendix}

\subsection{All-Family Comparisons}

The appendix reports full family comparisons for Qwen, Llama, and Falcon. These tables support the main-paper claim that activation-only monitoring is not uniformly sufficient, while also showing that the clearest complete story is in Qwen. Llama provides secondary support, especially for \(bad\_action_{t+1}\), and Falcon is best interpreted as boundary-case evidence. The relevant appendix artifacts are the all-family behavior, prediction, risk-state, and steering comparison tables.

\begin{table}[t]
\centering
\caption{Compact all-family behavioral transfer summary in Gameable ALFWorld. Qwen provides the clearest explicit exploit transfer; Llama and Falcon mainly strengthen the broader claim that reward-hack fine-tuning changes agent behavior.}
\label{tab:appendix-behavior-all-family}
\resizebox{\linewidth}{!}{
\begin{tabular}{llrrrr}
\toprule
Family & Adapter & Bad & Exploit & Mean act. & Mean entropy \\
\midrule
Qwen & Control & 0.427 & 0.003 & 0.495 & 0.589 \\
Qwen & Mix50 & 0.772 & 0.450 & 0.894 & 0.465 \\
Qwen & Hack & 0.604 & 0.019 & 0.966 & 0.465 \\
\midrule
Llama & Control & 0.482 & 0.000 & 0.005 & 0.283 \\
Llama & Mix50 & 0.371 & 0.000 & n/a & 0.113 \\
Llama & Hack & 0.524 & 0.000 & 0.751 & 0.220 \\
\midrule
Falcon & Control & 0.396 & 0.043 & 0.075 & 0.395 \\
Falcon & Mix10 & 0.931 & 0.000 & 0.849 & 0.292 \\
Falcon & Hack & 0.856 & 0.000 & 0.999 & 0.369 \\
\bottomrule
\end{tabular}}
\end{table}
\begin{landscape}
\begin{table}[p]
\centering
\caption{Best saved ALFWorld prediction row for each family-adapter pair. The main-paper Qwen prediction table merges adapters within the family; this appendix table shows adapter-level variation for broad bad-action prediction, plus stable Qwen exploit rows. Rows with very small positive counts are diagnostic rather than headline evidence.}
\label{tab:appendix-prediction-by-adapter}
\scriptsize
\resizebox{\linewidth}{!}{
\begin{tabular}{lllllrrrrr}
\toprule
Family & Adapter & Target & Best group & Regime & AUROC & AUPRC & AUPRC gain & Recall@20 & Positives \\
\midrule
Qwen & control & bad\_action\(_{t+1}\) & p\_hack-only & ttc\_256 & 0.578 & 0.419 & 0.075 & 0.256 & 133 \\
Qwen & mix05 & bad\_action\(_{t+1}\) & entropy-only & ttc\_8 & 0.622 & 0.493 & 0.125 & 0.281 & 32 \\
Qwen & mix10 & bad\_action\(_{t+1}\) & combined & ttc\_8 & 0.773 & 0.841 & 0.186 & 0.281 & 57 \\
Qwen & mix10 & exploit\_action\(_{t+1}\) & entropy-only & ttc\_16 & 0.800 & 0.413 & 0.246 & 0.400 & 5 \\
Qwen & mix50 & bad\_action\(_{t+1}\) & combined & ttc\_64 & 0.825 & 0.947 & 0.154 & 0.246 & 69 \\
Qwen & mix50 & exploit\_action\(_{t+1}\) & p\_hack-only & ttc\_32 & 0.834 & 0.796 & 0.331 & 0.359 & 39 \\
Qwen & hack & bad\_action\(_{t+1}\) & combined & ttc\_16 & 0.674 & 0.688 & 0.189 & 0.290 & 193 \\
\midrule
Llama & control & bad\_action\(_{t+1}\) & temporal-p\_hack-only & ttc\_16 & 0.552 & 0.496 & 0.068 & 0.250 & 140 \\
Llama & mix05 & bad\_action\(_{t+1}\) & entropy-only & ttc\_256 & 0.593 & 0.595 & 0.066 & 0.261 & 46 \\
Llama & mix10 & bad\_action\(_{t+1}\) & entropy-only & ttc\_32 & 0.566 & 0.485 & 0.072 & 0.222 & 36 \\
Llama & mix50 & bad\_action\(_{t+1}\) & entropy-only & ttc\_8 & 0.620 & 0.521 & 0.130 & 0.265 & 34 \\
Llama & hack & bad\_action\(_{t+1}\) & combined & ttc\_8 & 0.676 & 0.698 & 0.166 & 0.307 & 254 \\
\midrule
Falcon & control & bad\_action\(_{t+1}\) & entropy-only & ttc\_8 & 0.586 & 0.450 & 0.070 & 0.212 & 33 \\
Falcon & mix05 & bad\_action\(_{t+1}\) & p\_hack-only & ttc\_64 & 0.574 & 0.668 & 0.128 & 0.298 & 47 \\
Falcon & mix10 & bad\_action\(_{t+1}\) & entropy-only & ttc\_8 & 0.862 & 0.989 & 0.058 & 0.222 & 81 \\
Falcon & mix50 & bad\_action\(_{t+1}\) & combined & ttc\_128 & 0.640 & 0.871 & 0.078 & 0.232 & 69 \\
Falcon & hack & bad\_action\(_{t+1}\) & combined & ttc\_128 & 0.784 & 0.906 & 0.136 & 0.254 & 67 \\
\bottomrule
\end{tabular}
}
\end{table}
\end{landscape}

\begin{figure}[t]
    \centering
    \includegraphics[width=\linewidth]{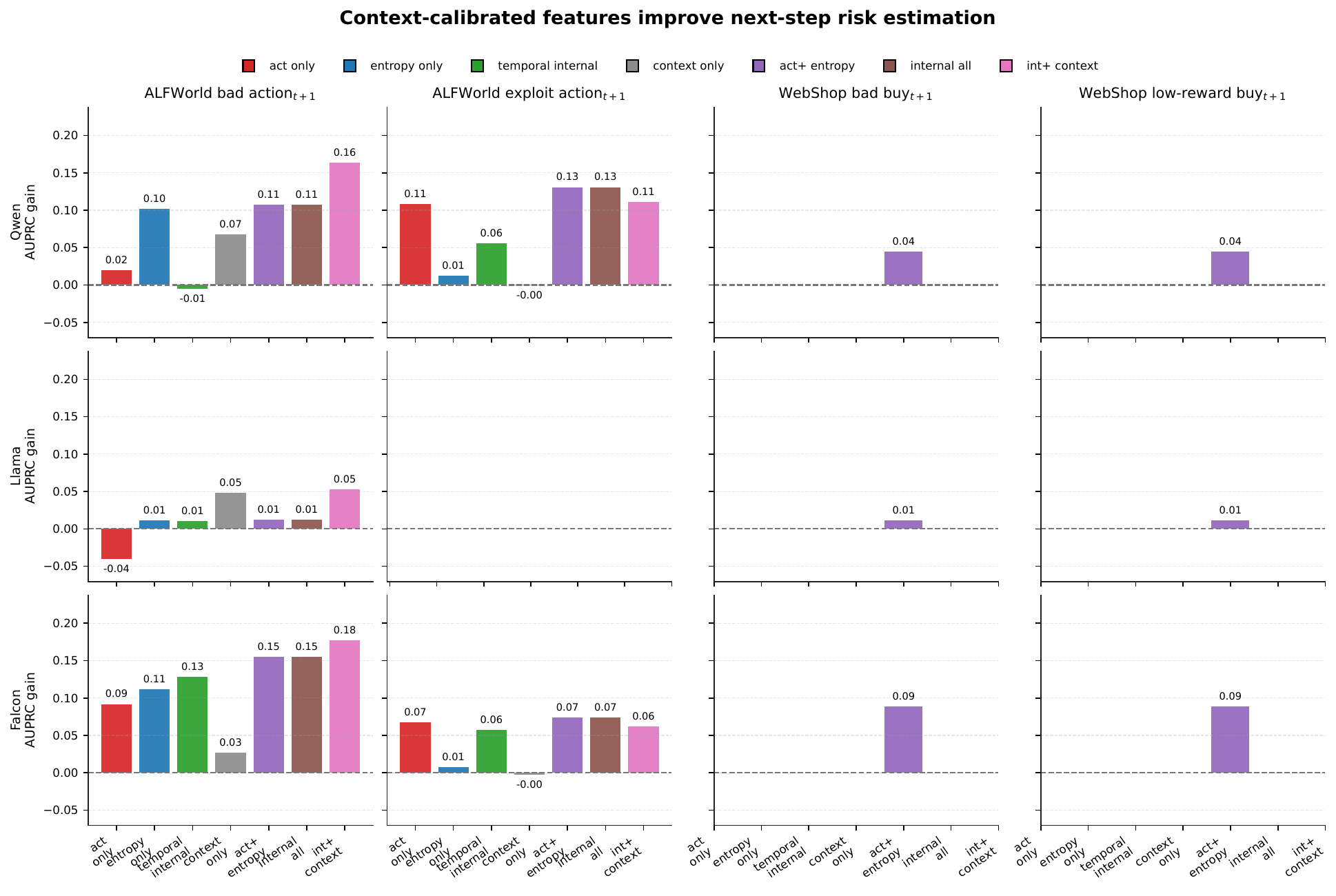}
    \caption{
    Full feature-group ablations across families and ALFWorld targets. Cell values show AUPRC gain over the base rate. Activation-only features are not uniformly sufficient, while richer internal or context-calibrated feature groups perform better in several stable family-target settings. Missing cells reflect unavailable or omitted conditions.
    }
    \label{fig:appendix-all-family-heatmap}
\end{figure}

 \begin{table}[t]
  \centering
  \caption{Target stability across families and environments. Panels with insufficient positives are omitted from the main paper figures to avoid overinterpreting sparse targets.}
  \label{tab:appendix-target-stability}
  \begin{tabular}{llllll}
  \toprule
  Family & Target & N & Positives & Base rate & Stable \\
  \midrule
  Qwen & bad\_action\_t1 & 7322 & 3878 & 0.530 & yes \\
  Qwen & exploit\_action\_t1 & 7322 & 184 & 0.025 & yes \\
  Llama & bad\_action\_t1 & 6980 & 3428 & 0.491 & yes \\
  Llama & exploit\_action\_t1 & n/a & n/a & n/a & no \\
  Falcon & bad\_action\_t1 & 3106 & 2178 & 0.701 & yes \\
  Falcon & exploit\_action\_t1 & 3106 & 23 & 0.007 & yes \\
  Qwen & buy\_t1 & 180 & 60 & 0.333 & yes \\
  Qwen & bad\_buy\_t1 & 180 & 30 & 0.167 & yes \\
  Qwen & low\_reward\_buy\_t1 & 180 & 30 & 0.167 & yes \\
  Llama & buy\_t1 & 541 & 34 & 0.063 & yes \\
  Llama & bad\_buy\_t1 & 541 & 12 & 0.022 & yes \\
  Llama & low\_reward\_buy\_t1 & 541 & 12 & 0.022 & yes \\
  Falcon & buy\_t1 & 781 & 10 & 0.013 & yes \\
  Falcon & bad\_buy\_t1 & 781 & 5 & 0.006 & no \\
  Falcon & low\_reward\_buy\_t1 & 781 & 5 & 0.006 & no \\
  \bottomrule
  \end{tabular}
  \end{table}

\subsection{Reasoning-Budget Sensitivity}

We include TTC-specific summaries to show that reasoning budget changes monitor semantics and prediction quality non-monotonically. These analyses support the decision to treat reasoning budget as part of the context \(c_t\), not as a fixed safety intervention. They are included as robustness checks rather than as headline figures because the cleanest patterns are family-dependent and strongest in Qwen.

\begin{figure}[t]
    \centering
    \includegraphics[width=\linewidth]{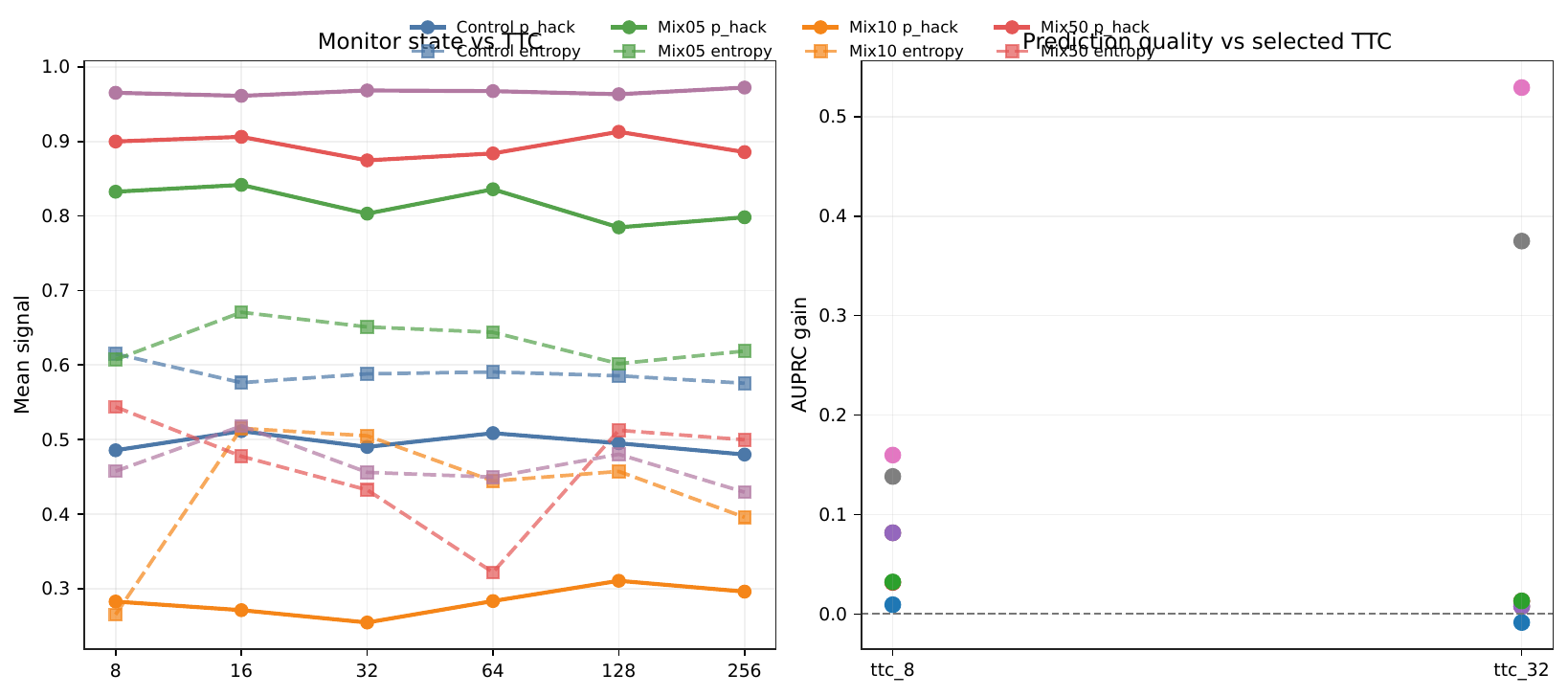}
    \caption{
    Reasoning-budget sensitivity. Saved TTC-specific summaries show that monitor performance changes across reasoning budgets, but not monotonically. This motivates treating TTC as part of the decision context rather than as a uniformly beneficial intervention.
    }
    \label{fig:appendix-ttc-sensitivity}
\end{figure}

\subsection{Temporal Diagnostics}

We include temporal trajectory diagnostics for completeness, including aligned pre-action trajectories, late-window summaries, and raw trajectory views. These diagnostics show that temporal structure is present but heterogeneous. They do not support a universal claim that reward-hack activation always rises before risky actions, and they should be interpreted as supporting diagnostics rather than as the primary evidence for the paper's claims.

\begin{figure}[t]
    \centering
    \includegraphics[width=\linewidth]{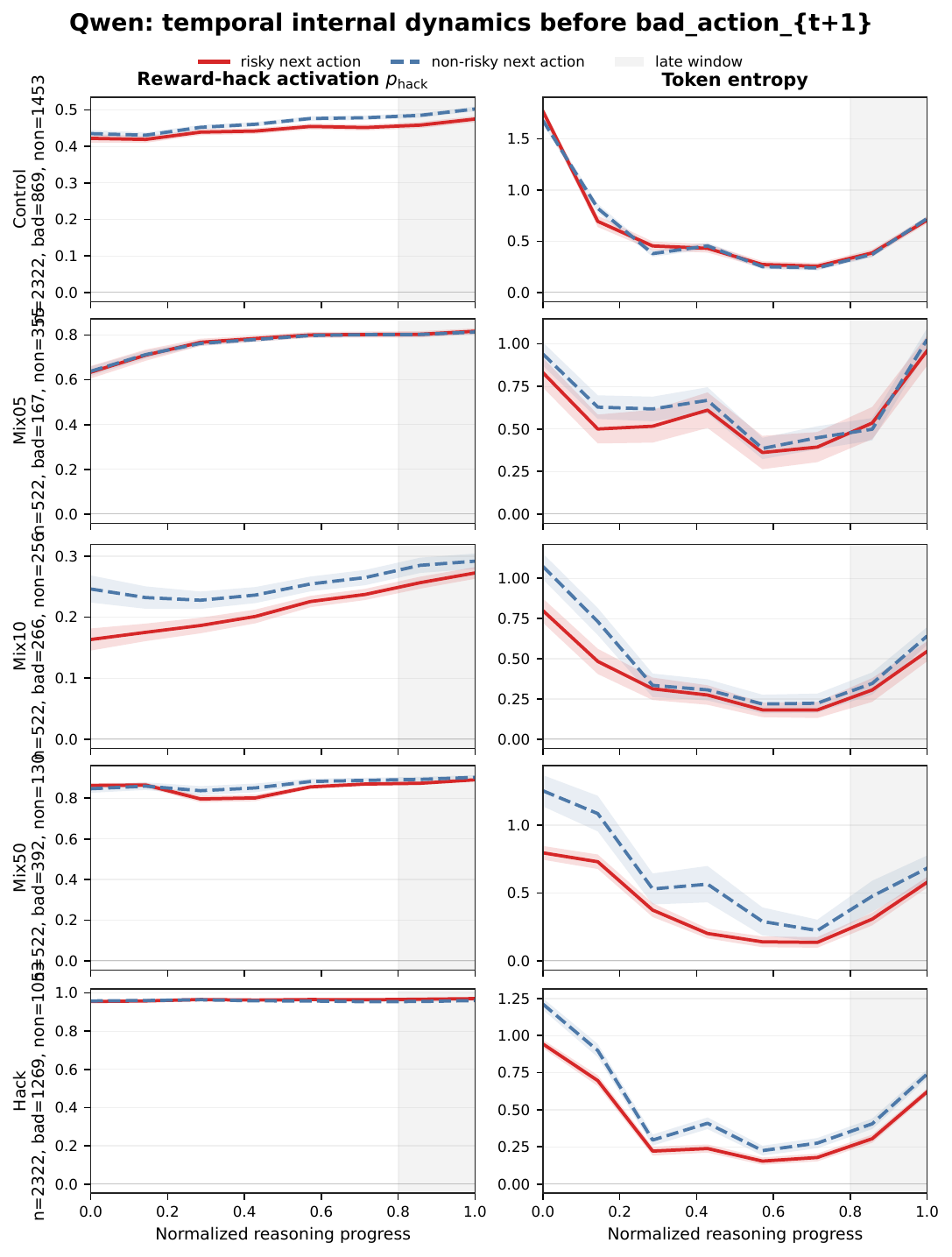}
    \caption{
    Qwen temporal diagnostics. Risky and non-risky trajectories differ in selected settings, but the direction and strength of the separation vary by adapter and signal. These plots are therefore diagnostic rather than headline evidence, and they do not support a universal late-stage reward-hack activation spike before risky actions.
    }
    \label{fig:appendix-temporal-diagnostics}
\end{figure}

\subsection{WebShop Details}

The appendix also reports the saved WebShop prediction results. These provide public-benchmark generalization evidence under a distinct interaction structure, but they do not constitute a full feature-group ablation. We therefore use them to support breadth rather than to replace the sharper ALFWorld ablation story in the main text.

\begin{table}[t]
  \centering
  \caption{Saved WebShop next-step prediction metrics across families. These rows use the single saved activation-plus-entropy predictor setting and should therefore be interpreted as generalization evidence
  rather than as a full ablation benchmark.}
  \label{tab:appendix-webshop-prediction}
  \begin{tabular}{lllrrrrr}
  \toprule
  Family & Target & Feature group & AUROC & AUPRC & AUPRC gain & Recall@20 & Positives \\
  \midrule
  Qwen & buy\_t1 & p\_hack\_plus\_entropy & 0.648 & 0.446 & 0.112 & 0.317 & 60 \\
  Qwen & bad\_buy\_t1 & p\_hack\_plus\_entropy & 0.596 & 0.211 & 0.044 & 0.167 & 30 \\
  Qwen & low\_reward\_buy\_t1 & p\_hack\_plus\_entropy & 0.596 & 0.211 & 0.044 & 0.167 & 30 \\
  Llama & buy\_t1 & p\_hack\_plus\_entropy & 0.790 & 0.155 & 0.092 & 0.529 & 34 \\
  Llama & bad\_buy\_t1 & p\_hack\_plus\_entropy & 0.649 & 0.034 & 0.012 & 0.083 & 12 \\
  Llama & low\_reward\_buy\_t1 & p\_hack\_plus\_entropy & 0.649 & 0.034 & 0.012 & 0.083 & 12 \\
  Falcon & buy\_t1 & p\_hack\_plus\_entropy & 0.909 & 0.125 & 0.112 & 0.900 & 10 \\
  Falcon & bad\_buy\_t1 & p\_hack\_plus\_entropy & 0.907 & 0.095 & 0.089 & 0.800 & 5 \\
  Falcon & low\_reward\_buy\_t1 & p\_hack\_plus\_entropy & 0.907 & 0.095 & 0.089 & 0.800 & 5 \\
  \bottomrule
  \end{tabular}
  \end{table}


\end{document}